\documentclass{article}

       \usepackage[final]{neurips_2020}

\usepackage[utf8]{inputenc} % allow utf-8 input
\usepackage[T1]{fontenc}    % use 8-bit T1 fonts
\usepackage[colorlinks]{hyperref}       % hyperlinks
\usepackage{url}            % simple URL typesetting
\usepackage{booktabs}       % professional-quality tables
\usepackage{amsfonts}       % blackboard math symbols
\usepackage{nicefrac}       % compact symbols for 1/2, etc.
\usepackage{microtype}      % microtypography
\usepackage{authblk}
\usepackage{graphicx}
\usepackage{multirow}
\usepackage{placeins}
\usepackage{amsmath}
\usepackage{bm}
\usepackage[dvipsnames]{xcolor}
\colorlet{LightRubineRed}{RubineRed!70!}
\colorlet{Mycolor1}{green!10!orange!90!}

\title{HINDI/BENGALI SENTIMENT ANALYSIS USING TRANSFER LEARNING AND JOINT DUAL INPUT LEARNING WITH SELF ATTENTION}

\author{
\begin{tabular}[t]{cc} 
  Shahrukh Khan  \hspace{4cm} Mahnoor Shahid\\
  \textbf{shkh00001@uni-saarland.de}  \hspace{2cm} \textbf{mash00001@uni-saarland.de} \\
\end{tabular}

\url{https://github.com/shahrukhx01/nnti_hindi_bengali_sentiment_analysis}
}

\begin{document}

\maketitle

\begin{abstract}
 Sentiment Analysis typically refers to using natural language processing, text analysis and computational linguistics to extract affect and emotion based information from text data. Our work explores how we can effectively use deep neural networks in transfer learning and joint dual input learning settings to effectively classify sentiments and detect hate speech in Hindi and Bengali data. We start by training Word2Vec word embeddings for Hindi \textbf{HASOC dataset} and Bengali hate speech and then train LSTM and subsequently, employ parameter sharing based transfer learning to Bengali sentiment classifiers by reusing and fine-tuning the trained weights of Hindi classifiers with both classifier being used as baseline in our study. Finally, we use BiLSTM with self attention in joint dual input learning setting where we train a single neural network on Hindi and Bengali dataset simultaneously using their respective embeddings.
\end{abstract}

\section{Introduction}

There have been certain huge breakthroughs in the field of Natural Language Processing paradigm with the advent of attention mechanism and its use in transformer sequence-sequence models coupled with different transfer learning techniques have quickly become state-of-the-art in multiple pervasive Natural Language Processing tasks such as classification, named entity recognition etc. In our work we reproduce some of that recent work related to the sentiment analysis and classification on \textbf{Hindi HASOC dataset} here we reproduce sub-task A which deals with whether a given tweet has hate-speech or not. Moreover, this also serves as a source domain in the subsequent transfer learning task, where, we take the learned knowledge from Hindi sentiment analysis domain to similar binary \textbf{Bengali sentiment analysis} task. \newline\newline
Given the similar nature of both Bengali and Hindi sentiment analysis tasks (i.e., binary classification), we conceptualized the problem as joint dual input learning setting on top of reproducing the work of  \hyperref[selfattention]{Zhouhan Lin et al., 2017} where they suggested how we can integrate self attention with BiLSTMs and have a matrix representation containing different aspects for each sequence which results in sentence embeddings whilst performing sentiment analysis and text classification more broadly. One significant beneficial side effect of using such approach is that the attention matrix can easily be visualized which imply we can see what portions of the sequence attention mechanism has put more impetus on via its generated summation weights, this visualization technique played pivotal role in selecting the number of attention hops $r$ also referred to as how many attention vectors of summation weights for each sequence in our study. Moreover, we employed this approach in a joint dual input learning setting where we have single neural network that is trained on Hindi and Bengali data simultaneously. 

\section{Baseline Methodology}
\subsection{Data Pre-processing}
\label{gen_inst}
\subsubsection{Hindi Data Pre-processing}
To begin with, we have done the pre-processing of the dataset where we have removed the punctuation, usernames and stop-words, along with normalizing the text data to lower case. Also we explored the distribution of response variable which is binary variable indication whether a given tweet contains hate speech or not. Following are the statistics for that variable: 
\begin{table}[!htb]
\begin{center}
  \caption{Tweets divided into Negative vs NonNegative Sentences}
  \centering
  \begin{tabular}{p{2cm}  l p{2cm} }
    \toprule                \
    Flag     & No. of Sentences     & Proportion \\
    \midrule
    HOF     & 2469   & 52.92\% \\
    NOT     & 2196   & 47.07\%  \\
    \bottomrule
  \end{tabular}
\end{center}
\end{table}
\paragraph{Handling emojis}
By analyzing the sentiment of the emojis we can draw several noteworthy conclusions which might be motivating to the study of interest. 
Hence, with the intention to obtain the emojis from the Hindi Dataset we have used regular expressions and have specified all the Emoji Unicode Blocks in the pattern and found just 2 emojis in the entire hindi dataset. \textit{So, removing them and not removing them will not make any difference in this setting.  }

\paragraph{Handling Hashtags}
We have used the matcher class from the Spacy package (python library) to match the sequences of the tokens, based on pattern rules and obtained the hash tags for both the negative and non-negative sentences. Mostly, there is great influence of hashtags on the sentences, as just by observing the words that are used as hashtags or by perceiving the high volumes of certain hashtags can direct us to the subject of the content or the trending topic. Correspondingly, it can affect the strength of the sentiment in a sentence, for example multiple negative hashtags can increase the negative sentiment of a tweet.

For these certain reasons we have settled not to eliminate them. 

\begin{figure}[!htb]
  \centering
  
  \includegraphics[width=10.2cm, height=5.1cm]{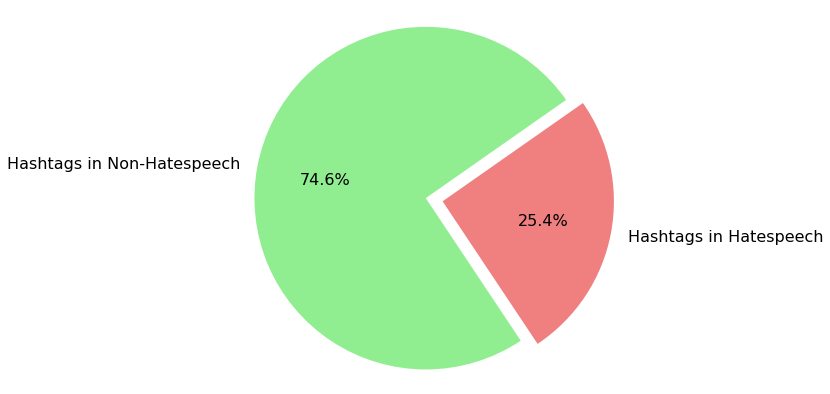}
%   \fbox{\rule[-.5cm]{0cm}{4cm} \rule[-.5cm]{4cm}{0cm}}
  \caption{Hashtags in Hindi Dataset for Negative vs Non-Negative Sentences based on label column}
\end{figure}

\begin{figure}[!htb]
  \centering
  \includegraphics[width=11cm, height=5.5cm]{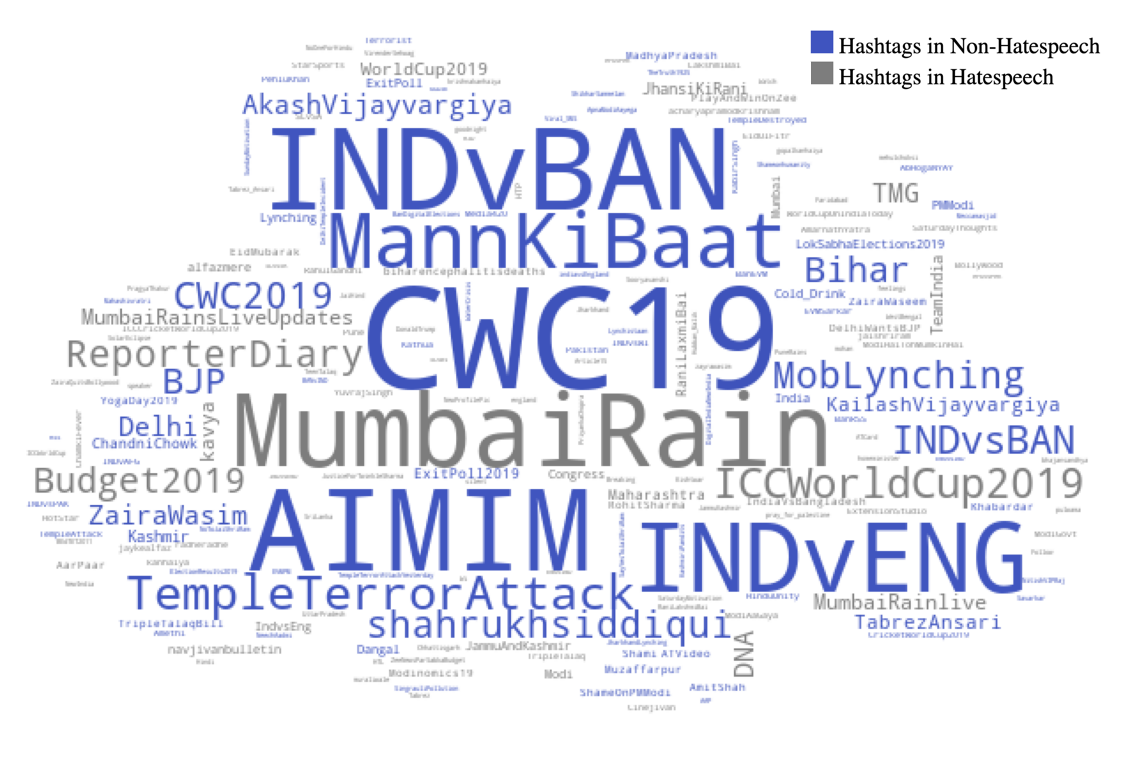}
%   \fbox{\rule[-.5cm]{0cm}{4cm} \rule[-.5cm]{4cm}{0cm}}
  \caption{WordCloud indicating the most frequent hashtags in the Hindi Dataset}
\end{figure}

\subsubsection{Bengali Data Pre-processing}
We started by first splitting the actual Bengali dataset and taking 2500 positive label examples, and 2500 negative label examples sampled randomly from the actual dataset of 30000 examples. This roughly makes the Bengali subset equivalent to the Hindi dataset and also avoids the problem of class imbalance since we have equal number of examples for both classes. \newline\newline
We performed a short analysis on the Bengali dataset as well, for emojis and hashtags, using the same approach as mentioned in section 3. The results we obtained were 27 hashtags with only 4 of them being in the negative sentences and got 7426 emojis of various kind with 5241 emojis used in the hatespeech texts and 2185 in non-hatespeech texts. The most frequently used top 6 emojis in the Bengali dataset can be observed from Figure 3.

\begin{figure}[!htb]
  \centering
  \includegraphics[width=14cm, height=6cm]{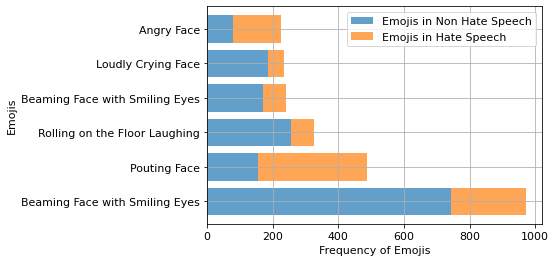}
%   \fbox{\rule[-.5cm]{0cm}{4cm} \rule[-.5cm]{4cm}{0cm}}
  \caption{Frequency of Emojis used in the Sentences of Bengali Dataset}
\end{figure}
\FloatBarrier
As it is apparent from the above stacked chart each type of emoji has more volume in either sentiment class, which can help the classifier in classification of the sentiment. So, for the very same reason, we have not deviated from the text pre-processing pipeline adapted in the Hindi dataset, also, we retained both the emojis and hashtags here as well for the very same reasons as described earlier.

\subsection{Word Embeddings}
Starting with the Hindi dataset, we pre-processed the dataset as per the pipeline described in section 2.1.1. We prepared the training dataset in which employed sub-sampling technique in which we first computed the probability of keeping the word using the following formula:
\[ P_{keep}(w_i) = (\sqrt{\dfrac{z(w_i)}{0.000001}} + 1) . \dfrac{0.000001}{z(w_i)} \]
Where $z(w_i)$ is the relative frequency of th word in the corpus. Hence we used $P_{keep}(w_i)$ for each context word while sampling context words for a given word and randomly dropped frequent context words by comparing them against a random threshold sampled each time from uniform distribution, since if we kept all the frequent words in our context for training data, we may not get rich semantic relationship between the domain specific words since frequent words like "the", "me" etc don't necessarily carry much semantic meaning in a given sequence. Hence dropping randomly dropping them made more sense as compared to keeping or dropping all of them. Also, another important design decision that we made here was to curate the train set for Word2Vec only once before training the model as opposed to creating a different one for each epoch as we were randomly sub-sampling context words, because the earlier mentioned approach gives faster execution time for training the model while the model also converged well to a relatively low train loss value as well. Furthermore, for choosing hyper-parameters we performed the following analysis.

\begin{figure}[!htb]
  \centering
  \includegraphics[width=13cm, height=6.7cm]{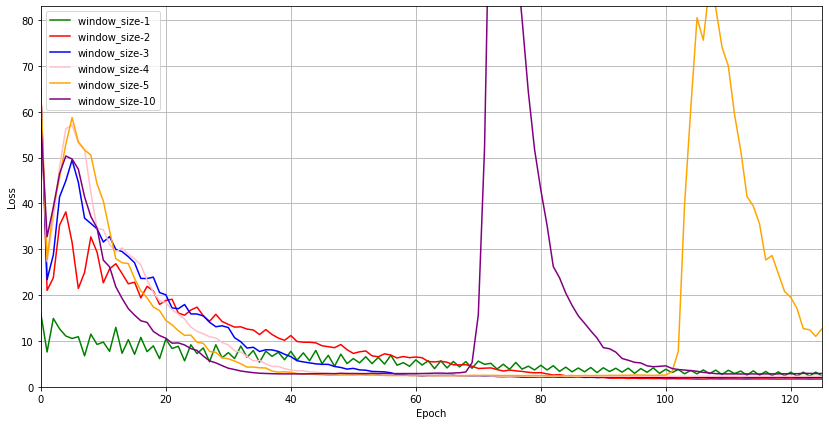}
%   \fbox{\rule[-.5cm]{0cm}{4cm} \rule[-.5cm]{4cm}{0cm}}
  \caption{Train loss convergence for different values of window size with fixed embedded size = 300 Hindi Dataset}
\end{figure}
As it is apparent from the above visualization WordVec models with smaller context windows converged faster and had better train loss at the end of training process. However, in order to retain some context based information we selected the window size 2 as it has contextual information as well the model had better train loss.\newline
\begin{table}[!htb]
\begin{center}
  \caption{Different combinations of hyper parameters from Hindi Dataset}
  \centering
  \begin{tabular}{c|ccc}
    \toprule    
    
    Embedded Size & Learning Rate  & Window Size  & Min Loss Score\\
    \midrule
    \multirow{9}{*}{300} & \multirow{6}{*}{0.05} & 1 & 0.841\\
    & & 2  & 1.559  \\
    & &3 & 1.942 \\
    & &4 & 2.151\\
    & &5  & 2.321\\
    & &10 & 2.792\\
    \cmidrule(r){2-4}
    & \multirow{3}{*}{0.01} & 1 & 1.298 \\
    & & 2 & 3.295 \\
    & & 10 & 2.747 \\
    \cmidrule(r){2-4}
    & \multirow{3}{*}{0.1} & 1 & 1.311 \\
    & & 2 & 1.557 \\
    & & 10 & 3.551 \\

    \bottomrule
  \end{tabular}
\end{center}
\end{table}
\FloatBarrier
After testing different values for hyper-parameters with different combinations, this was observed that for the better performance of the model, they should be set to \textbf{Epochs = 500, Window Size = 2, Embedded Size = 300, and Learning Rate = 0.05} in the case of our study.
\newline
Also, we have set \textbf{Cross Entropy Loss} as the criterion used for adjusting the weights during the training phase. When softmax converts logits into probabilities then, Cross-Entropy takes those output probabilities (p) and measures the distance from the truth values to estimate the loss. Cross entropy loss inherently combines \textbf{log softmax and negative log likelihood loss} so we didn't apply log softmax on the output of our Word2Vec model.

For optimization we have selected \textbf{Adam (Adaptive Moment Estimation algorithm)} which is an optimization technique that, at present, is very much recommended for its computational efficency, low memory requirement, invariant to diagonal rescale of the gradients and extremely better results for problems that are large in terms of data/parameters or for problems with sparse gradients. Adam provides us with the combination of best properties from both AdaGrad and RMSProp, and is often used as an alternative for SGD + Nesterov Momentum as proposed by \hyperref[adam]{Diederik P. et al., 2015}.
\subsection{Baseline Models}
For the choice of baseline we reproduced the work by \hyperref[lstmembed]{Jenq-Haur Wang et al., 2018} which primarily focuses on performing sentiment classification on short social media texts using long short-term memory neural networks using distributed representations of Word2Vec learned using Skip-gram approach. We chose to reproduce their work for our baseline as they also were using Word2Vec Skip-gram based distributed representation of words and also since our datasets were also sourced from social media. Moreover, the neural network LSTM is an upgraded variant of the RNN model, that serves as the remedy to some extent of the problems that requires learning long-term temporal dependencies; due to vanishing gradients, since LSTM uses gate mechanism and memory cell to control the memorizing process. 

\subsubsection{Hindi Neural Sentiment Classifier Baseline}
Firstly, we applied the same text pre-processing pipeline as described in 2.2.1. Then we divided the Hindi dataset into 2985 examples in train set, 746 examples in validation set and 932 examples in test set. We then implemented the architecture for LSTM classifier which used pre-trained 300 dimensional word embeddings obtained as described in section 2.2. We used Adam optimizer for the same reasons listed in section 2.2 with the initial learning rate of $10^{-4}$ which helped the train and validation loss to converge at a relatively fast rate, the optimizer didn't optimize the weights of embedding layer via gradient optimization since they were pre-trained already. Moreover, we chose binary cross entropy loss function as we are doing binary classification. Binary cross entropy is designed to work with a single sigmoid function as output activation, which we have included in our network, In model architecture we used $8$ layers of LSTMs with each having hidden dimension of 64 followed by a dropout layer with dropout probability of $0.5$ to counterbalance over fitting and finally fully connected output layer wrapped by a sigmoid activation function since our target is binary and sigmoid is the ideal choice for binary classification given its mathematical properties. We kept a batch size of $32$ and trained the model for $30$ epochs while monitoring its accuracy and loss on validation set. The choice of hyper-parameters was made after trying different combinations and we chose the bet set of hyper-parameters while monitoring the validation set accuracy.

\subsubsection{Bengali Neural Transfer Learning Based Sentiment Classifier Baseline}
Firstly, we applied the same text pre-processing pipeline as described in 2.2.2. Then we divided the Bengali dataset into 3194 examples in train set, 798 examples in validation set and 998 examples in test set. Similarly to the Hindi sentiment classification pipeline, we first obtained the word embeddings for Bengali data using the Word2Vec skip-gram approach, the same set of hyper-parameters that we chose for Hindi dataset, worked fine here well, so we didn't tune the hyper-parameters here, as the model's train loss converged to similar value we had for the Hinidi dataset. Subsequently, we then same the architecture for LSTM based classifier architecture as explained in 2.2.2. Since our goal here was to perform transfer learning and re-use and fine-tune the learned weights of Hindi classifier. We replaced the Hindi embeddings layer with Bengali 300 dimensional embedding layer and also didn't optimize its weights during training. The loaded the weights from Hindi classifier for LSTM layers and fully connected layer to apply parameter sharing based task specific transfer learning. Additionally, we trained the Bengali classifier for 30 epochs with batch size of 32 and using the Adam optimizer with initial learning rate $10^{-4}$ while using binary cross entropy function for computing loss on training and validation set. The choice of batch size hyper-parameter was made after trying different values and we chose the best hyper-parameter while monitoring the validation set accuracy. After training the classifier using the pre-trained weights from Hindi classifier, we got better performance results to the Hindi baseline, this implies task based transfer learning actually boosted the performance of Bengali classifier and it performed better.\newline \newline

\section{Our Work}
The LSTM based classifier coupled with transfer learning in Bengali domain do a fairly good job for providing the baselines in our study. However, one main prominent shortcoming of Recurrent Neural Networks based architectures is they fall short to capture the dependencies between words that are too distant from each other. LSTM's forget gate enables it to retain information of the historical words in the sequence however, it doesn't completely resolve the RNN based networks vanishing gradients problem. We wanted to investigate whether using self attention with LSTMs would improve our model's performance. Also, we propose the joint dual input learning setting where both Hindi and Bengali classification tasks can benefit from each other rather than the transfer learning setting where only the target task takes the advantage of pre-training.
\subsection{Hindi \& Bengali Self Attention Based Joint Dual Input Learning BiLSTM Classifier}
Instead of training two separate neural networks for Hindi \& Bengali, here we  simultaneously trained a joint neural network with the same architecture on Hindi and Bengali data in parallel, and optimized its weights using the combined binary cross entropy loss over Hindi \& Bengali datasets respectively, we also added the Hindi and Bengali batches' attention loss to the joint loss in order to avoid overfitting, which we would present in detail in the subsequent sections. Here we switched between the embedding layers based on the language of the batch data. Following is the block architecture we propose.
\begin{figure}[!htb]
\caption{\textbf{Hindi Bengali Joint Dual Input Learning Architecture}}
  \centering
  \includegraphics[width=12cm, height=9cm]{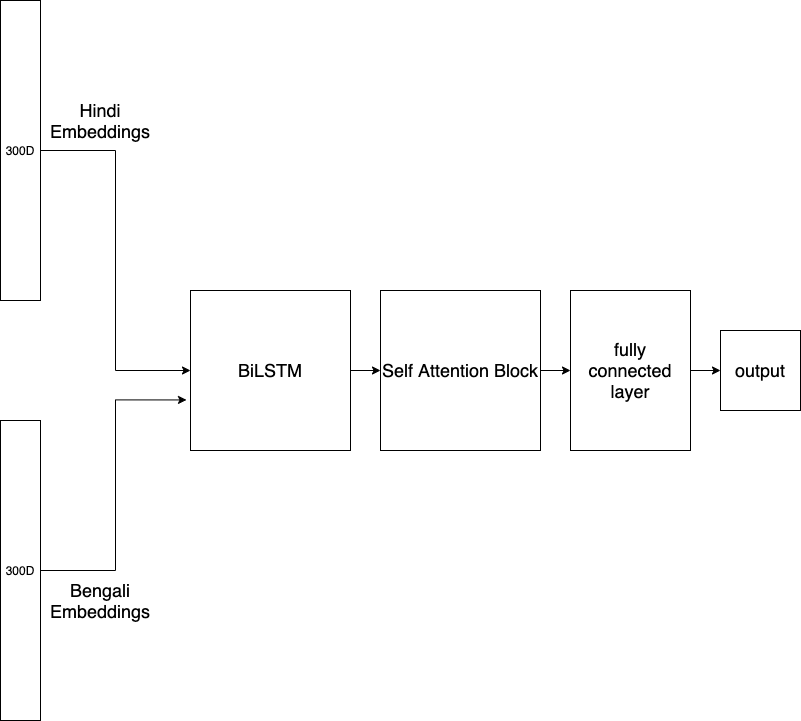}
%   \fbox{\rule[-.5cm]{0cm}{4cm} \rule[-.5cm]{4cm}{0cm}}
\end{figure}
\FloatBarrier
One major benefit of using such technique is that it increases model capability of generalization, since the size of training data set roughly doubles given if both languages have equal number of training examples.
Consequently, it reduces the risk of over-fitting.

We started with re-producing the work of \hyperref[selfattention]{Zhouhan Lin et al., 2017} where they proposed the method of \textbf{"A Structured Self-attentive Sentence Embedding"} on Hindi dataset. The key idea of that work was to propose document level embeddings by connecting self attention mechanism right after a Bi-directional LSTM, which leverages information of both past and future in the sequence as opposed to unidirectional LSTM which only relies on past information in the sequence. The self attention mechanism results in a matrix of attention vectors which are then used to produce sentence embeddings, each of them equivalent to the length of the sequence and number of vectors depends on the value of $r$ which is the output dimension of the self attention mechanism, where each vector is representing how attention mechanism is putting more relative weight on different tokens in the sequence. Following are the key takeaways how self attentive document embeddings are produced: \newline
We start with a input text T of $(n, d)$ dimension, where n are the number of tokens, each token is represented by its embedding $e$ in the sequence and $d$ is the embedding dimension.
\[ T = [e_1,e_2,e_3,...,e_n] \]
\newline\newline
Token embeddings are then fed into the BiLSTM, which individually processes each token from left to right and left to right direction, each BiLSTM cell/layer producing two vectors of hidden states equivalent to length of sequence.

\[  [[\overrightarrow{h_1},\overrightarrow{h_2},....,\overrightarrow{h_n}],[\overleftarrow{h_1},\overleftarrow{h_2},....,\overleftarrow{h_n}]] = BiLSTM([e_1,e_2,e_3,...,e_n];\theta) \]
Here \textbf{H} is the concatenated form of bi-directional hidden states. If there are $l$ LSTM layers/cells then the dimension of \textbf{H} is going to be $(n, 2l)$. 
\[  \textbf{H}=[[\overrightarrow{h_1},\overrightarrow{h_2},....,\overrightarrow{h_n}],[\overleftarrow{h_1},\overleftarrow{h_2},....,\overleftarrow{h_n}]]\]

For self attention \hyperref[selfattention]{Zhouhan Lin et al., 2017} proposed having two weight matrices, namely $W_{s_1}$ with dimension $(d_a, 2l)$ and $W_{s_2}$ with dimension $(r, d_a)$, here $d_a$ is the hidden dimension of self attention mechanism and as described earlier $r$ is the number of attention vectors for a given text input and then we apply following set of operations to produce the attention matrix for input text T.
\[H_a = tanh(W_{s_1}H^T)\]
Here $H_a$ has dimensions $(d_a, n)$
\[A = softmax(W_{s_2}H_a)\]
Finally, we compute sentence/document level embeddings
\[M = AH\]
$A$ has dimensions $(r, n)$ and $M$ has dimensions $(r, 2l)$ and also, earlier the softmax applied along second dimension of $A$ normalizes attention weights so they sum up 1 for each attention vector of length $n$.\newline\newline The above work also proposed penalization term in place of regularization to counterbalance redundancy in embedding matrix $M$ when attention mechanism results in same summation weights for all $r$ hops, additionally, We initially started by setting this penalization term to $0.0$ however, as self-attention generally works well for finding long term dependencies the neural network started to overfit after few epochs of training on train data. 

We started with the same hyper-parameters setting of self attention block as described by \hyperref[selfattention]{Zhouhan Lin et al., 2017} while setting $r=30$ however, we started with no penalization to start with and found the best values for them while monitoring the validation set accuracy which are hidden dimension of $300$ for self attention, with $8$ layers of BiLSTM with hidden dimension of $32$ and also, the output of self attention mechanism (sentence embeddings $M$) goes into a fully connected layer with its hidden dimension set to $2000$, finally we feed the fully connected layer's results to output layer wrapped with sigmoid activation. The choice of loss function, learning rate and optimizer remains unchanged from the baseline, number of epochs are $20$ here. After training the model with hyper parameters suggested in the above text, we observed the model started to overfit on train data after few epochs and almost achieved $99\%$ train accuracy and loss less than $0.5$ average epoch train loss, in order to add the remedy for this we we visually inspected the few of the examples from test set in attention matrix with confidence $> 0.90$ and observed for longer sequences the attention mechanism worked as expected however, as the sequence length decreased the attention mechanism started producing roughly equal summation weights on all $r$ hops which intuitively makes since in short sequences all tokens would carry more semantic information however, this result in redundancy in attention matrix $A$ and in embedding matrix $M$. Below we present some of the examples from Hindi test set, also since showing all the vectors would make it redundant so we only present $5$ vectors for a given sequence even though we had $r$ set to $30$ which implies we had 30 vectors for each sequence.
\begin{figure}[!htb]
\caption{\textbf{Attention vectors for a relatively longer Hindi sequence}}
  \centering
  \includegraphics[width=13cm, height=6.7cm]{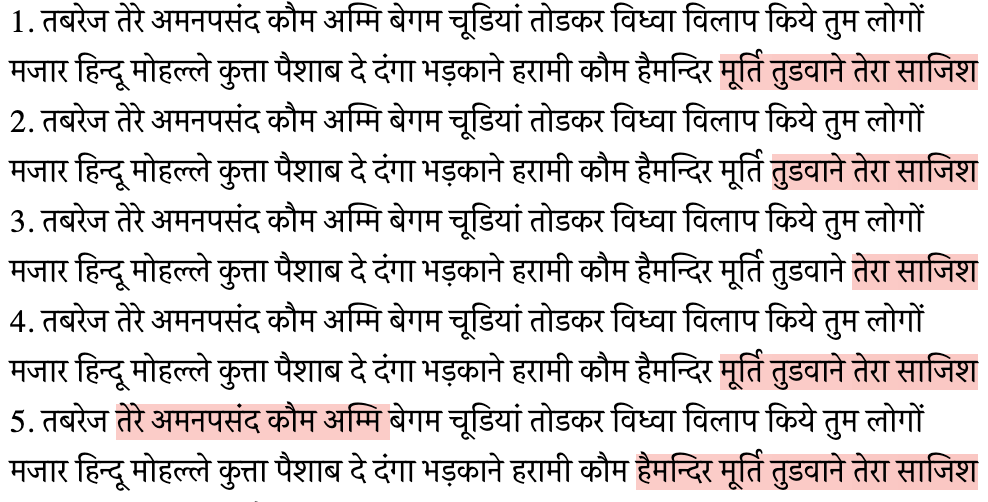}
%   \fbox{\rule[-.5cm]{0cm}{4cm} \rule[-.5cm]{4cm}{0cm}}
\end{figure}

\begin{figure}[!htb]
\caption{\textbf{Attention vectors for a relatively medium Hindi sequence}}
  \centering
  \includegraphics[width=13cm, height=6.7cm]{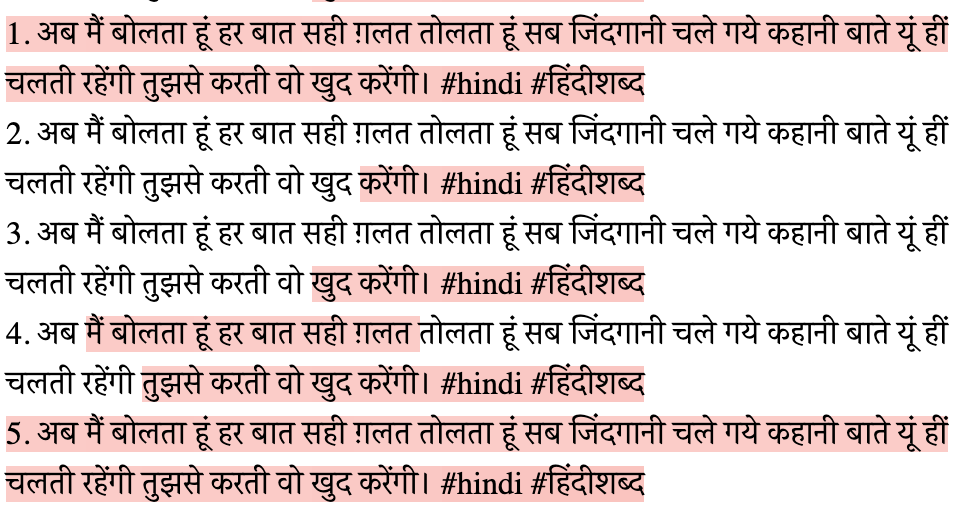}
%   \fbox{\rule[-.5cm]{0cm}{4cm} \rule[-.5cm]{4cm}{0cm}}
\end{figure}

\begin{figure}[!htb]
\caption{\textbf{Attention vectors for a short Hindi sequence}}
  \centering
  \includegraphics[width=6cm, height=4cm]{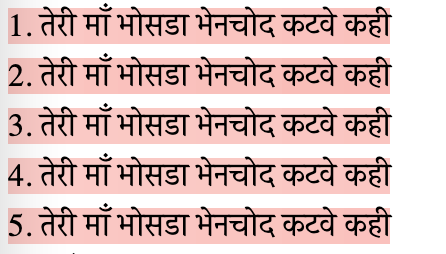}
%   \fbox{\rule[-.5cm]{0cm}{4cm} \rule[-.5cm]{4cm}{0cm}}
\end{figure}
\FloatBarrier
 Also, we performed the same analysis as we performed for Hindi data. Following we would also show few similar examples as we showed for Hindi sequences.
\begin{figure}[!htb]
\caption{\textbf{Attention vectors for a short Bengali sequence}}
  \centering
  \includegraphics[width=8cm, height=3cm]{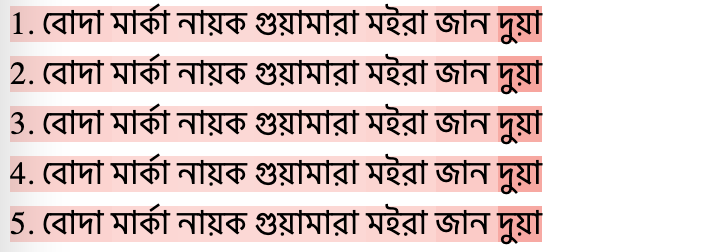}
%   \fbox{\rule[-.5cm]{0cm}{4cm} \rule[-.5cm]{4cm}{0cm}}
\end{figure}
\begin{figure}[!htb]
\caption{\textbf{Attention vectors for a short Bengali sequence}}
  \centering
  \includegraphics[width=8cm, height=3cm]{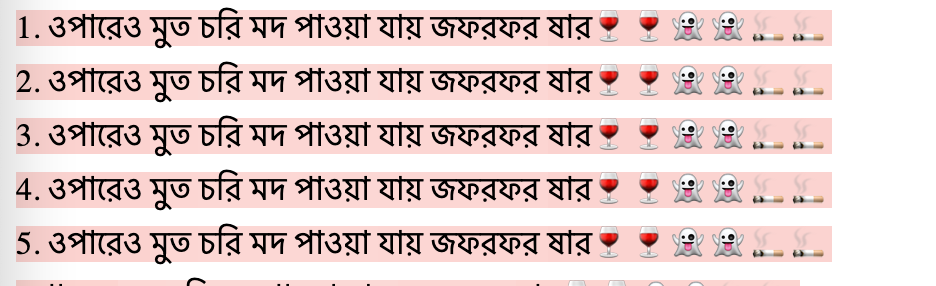}
%   \fbox{\rule[-.5cm]{0cm}{4cm} \rule[-.5cm]{4cm}{0cm}}
\end{figure}
\FloatBarrier
In order to counterbalance this redundancy we started increasing the value of penalization coefficient of attention mechanism in order to reduce the redundancy among the attention matrix and found penalization coefficient of $0.6$ produced the best validation set accuracy, similarly, the other form of diagnosis we performed was to actually reduce the number of attention hops , i.e., varying the hyper-parameter r and observed network with r = 20 had better performance on validation, alongside setting hidden size of attention mechanism to $150$ set as compared to r = 30 and hidden size =200 as suggested in the original work. Also, in order to avoid any over-fitting during in the BiLSTM block we used dropout in BiLSTM layers with a value of p = 0.5.

\section{Results}
\textbf{Note: \textcolor{LightRubineRed}{ Model names have links to respective model notebook. Precision, Recall and F-1 are macro averaged. SA refers to self attention. PRET refers to using pre-trained Hindi weights for the corresponding architecture and JDIL refers to joint dual input learning.}}
\begin{table}[!htb]
\begin{center}
  \caption{Results of evaluating the binary neural Hindi and Bengali sentiment classifiers on their respective test sets}
  \centering
  \begin{tabular}{lllll }
    \toprule                
    Model	&Accuracy	&Precision	&Recall	&F-1 Score     \\
    \midrule 
    \href{https://github.com/shahrukhx01/nnti_hindi_bengali_sentiment_analysis/blob/main/src/task2/1_hindi_lstm/Hindi_Binary_Classifier.ipynb}{LSTM-Hindi}	& 0.74 &	0.74&	0.74&	0.74\\
    \href{https://github.com/shahrukhx01/nnti_hindi_bengali_sentiment_analysis/blob/main/src/task2/2_bengali_lstm_pret/Bengali_Transfer_Learning_Binary_Classifier.ipynb}{LSTM-Bengali + PRET}	&0.77	&0.77	&0.77&	0.77\\
    \href{https://github.com/shahrukhx01/nnti_hindi_bengali_sentiment_analysis/blob/main/src/task3/1_hindi_bengali_bilstm_sa_jdil/Sentiment_Net.ipynb}{BiLSTM-Hindi/Bengali SA + JDIL (lang=Hindi)}	& \textbf{0.76}&  \textbf{0.76}&      \textbf{0.76}&     \textbf{0.76}\\
    \href{https://github.com/shahrukhx01/nnti_hindi_bengali_sentiment_analysis/blob/main/src/task3/1_hindi_bengali_bilstm_sa_jdil/Sentiment_Net.ipynb}{BiLSTM-Hindi/Bengali SA + JDIL (lang=Bengali)}	& \textbf{0.78}&  \textbf{0.78}&      \textbf{0.78}&     \textbf{0.78} \\
    \bottomrule
  
  \end{tabular}
\end{center}
\end{table}
\FloatBarrier

\section{Conclusion}

In our study we investigated whether self attention can enhance significantly the performance over uni-directional LSTM in the binary classification task setting, moreover, we also investigated when the tasks are same in our case binary classification in Hindi and Bengali language, whether how does transfer learning and joint dual input learning setting perform. Firstly we found when the lengths of sequences are not that long LSTMs can perform almost as good as using self attention since there are no very distant dependencies in sequences in most of the cases. Secondly, we observed that transfer learning in case similar or same tasks can be beneficial way of increasing the performance of target task which in our case was Bengali binary classification. However, by introducing the joint learning setting where we trained a single network for both task the Hindi classification task that was source task in transfer learning setting, also got benefited in joint learning setting as its performance improved. Moreover, such architecture provides implicit mechanism to avoid overfitting as it roughly doubled the dataset size when we trained a single network. Lastly, although self attention based mechanism improved our model's performance slightly, however, the performance gains were not significant one possible reason behind that could be since the input sequences weren't really long such as compared to Wikipedia article, or an online news article etc, so LSTM based model performed effectively as well because of the absence of very distant dependencies in the input sequences. In conclusion in such cases vanilla LSTMs should be the first choice as per \textbf{Occam Razor's principle} which suggests simpler models have less chances of overfitting on train data and better capability of generalization as compared to more complex models.

\section*{References}
\medskip

\small

[1] \label{lstmembed}An LSTM Approach to Short Text Sentiment Classification with Word Embeddings. Jenq-Haur Wang, Ting-Wei Liu, Xiong Luo and Long Wang. The 2018 Conference on Computational Linguistics and Speech Processing ROCLING 2018, pp. 214-223. https://www.aclweb.org/anthology/O18-1021.pdf

[2] \label{selfattention}A Structured Self-attentive Sentence Embedding. Zhouhan Lin and Minwei Feng, Cicero Nogueira dos Santos, Mo Yu,  Bing Xiang, Bowen Zhou, and Yoshua Bengio. Published as a conference paper at ICLR 2017. https://arxiv.org/pdf/1703.03130.pdf

[3] \label{ascpecttransfer}Exploiting Document Knowledge for Aspect-level Sentiment Classification. Ruidan He, Wee Sun Lee, Hwee Tou Ng, and Daniel Dahlmeier. https://arxiv.org/pdf/1806.04346.pdf

[4] Overview of the hasoc track at fire 2019: Hate speech and offensive content identification in indo-european languages. Thomas Mandl, Sandip Modha, Prasenjit Majumder, Daksh Patel, Mohana Dave, Chintak Mandlia, and Aditya Patel. In Proceedings of the 11th Forum for Information Retrieval Evaluation, pages 14–17, 2019.

[5] \label{adam}Adam: A Method for Stochastic Optimization
Diederik P. Kingma, Jimmy Ba. Published as a conference paper at ICLR 2015. https://arxiv.org/pdf/1412.6980.pdf

[6] Long Short-Term Memory. Jurgen Schmidhuber. Sepp Hochreiter. Neural Computation 9(8): 1735-1780, 1997. http://www.bioinf.jku.at/publications/older/2604.pdf

\end{document}